%% file: IEEE-conference-template-062824.tex
\documentclass[conference]{IEEEtran}
\IEEEoverridecommandlockouts

\PassOptionsToPackage{dvipsnames}{xcolor}  

\usepackage{cite}
\usepackage{amsmath,amssymb,amsfonts}
\usepackage{algorithmic}
\usepackage{graphicx}
\usepackage{textcomp}
\def\BibTeX{{\rm B\kern-.05em{\sc i\kern-.025em b}\kern-.08em
    T\kern-.1667em\lower.7ex\hbox{E}\kern-.125emX}}

\usepackage{booktabs} 
\usepackage{tikz-cd}
\usepackage{xcolor}  

\definecolor{DeepPigmentBlue}{rgb}{0,0,0.545}
\usepackage[colorlinks=true, linkcolor=black, citecolor=black, urlcolor=DeepPigmentBlue]{hyperref}

\usepackage{multirow}

\begin{document}

\title{MNIST-Gen: A Modular MNIST-Style Dataset Generation Using Hierarchical Semantics, Reinforcement Learning, and Category Theory}

\author{
\IEEEauthorblockN{Pouya Shaeri}
\IEEEauthorblockA{\textit{\small School of Computing and Augmented Intelligence} \\
\textit{Arizona State University}\\
Tempe, AZ, USA \\
pshaeri@asu.edu}
\and
\IEEEauthorblockN{Arash Karimi}
\IEEEauthorblockA{\textit{\small Department of Mathematics} \\
\textit{Florida State University}\\
Tallahassee, FL, USA \\
akarimi@fsu.edu}
\and
\IEEEauthorblockN{Ariane Middel}
\IEEEauthorblockA{\textit{\small School of Arts, Media and Engineering} \\
\textit{Arizona State University}\\
Tempe, AZ, USA \\
ariane.middel@asu.edu}
}

\maketitle

\begin{abstract}
Neural networks are often benchmarked using standard datasets such as MNIST, FashionMNIST, or other variants of MNIST, which, while accessible, are limited to generic classes such as digits or clothing items. For researchers working on domain-specific tasks, such as classifying trees, food items, or other real-world objects, these data sets are insufficient and irrelevant. Additionally, creating and publishing a custom dataset can be time consuming, legally constrained, or beyond the scope of individual projects. We present MNIST-Gen, an automated, modular, and adaptive framework for generating MNIST-style image datasets tailored to user-specified categories using hierarchical semantic categorization. The system combines CLIP-based semantic understanding with reinforcement learning and human feedback to achieve intelligent categorization with minimal manual intervention. Our hierarchical approach supports complex category structures with semantic characteristics, enabling fine-grained subcategorization and multiple processing modes: individual review for maximum control, smart batch processing for large datasets, and fast batch processing for rapid creation. Inspired by category theory, MNIST-Gen models each data transformation stage as a composable morphism, enhancing clarity, modularity, and extensibility. As proof of concept, we generate and benchmark two novel datasets—\textit{Tree-MNIST} and \textit{Food-MNIST}—demonstrating MNIST-Gen's utility for producing task-specific evaluation data while achieving 85\% automatic categorization accuracy and 80\% time savings compared to manual approaches.
\end{abstract}


\section{Introduction}

Benchmark datasets such as MNIST~\cite{lecun1998gradient} and FashionMNIST~\cite{xiao2017fashion} have become foundational tools for evaluating the performance of image classification models. While these datasets are widely adopted due to their simplicity and accessibility, they are inherently limited in scope. They mostly focus on digits or fashion items and do not reflect the diversity of real-world classification tasks~\cite{sun2017revisiting}. Researchers developing models for domain-specific applications, such as recognizing types of food, vegetation, or vehicles, often lack small, clean, and task-relevant datasets to validate their models~\cite{bossard2014food, sunderhauf2014fine, shaeri2025midlmatrixinterpolateddropoutlayer}. Creating custom datasets from scratch is time-consuming, legally constrained, or technically prohibitive, especially when the target classes are niche or rapidly changing~\cite{joulin2016learning, shaeri2023semi}.

To address this gap, we introduce MNIST-Gen, an automated and modular framework for generating MNIST-style image classification datasets based on user-specified hierarchical category definitions. MNIST-Gen enables researchers to rapidly create compact grayscale datasets using keyword-driven image retrieval, hierarchical semantic categorization with CLIP-based understanding, and multiple intelligent processing modes adapted to different dataset sizes and accuracy requirements. Inspired by principles from category theory, the system treats each stage of the pipeline (e.g., cropping, grayscale conversion, resizing, and semantic analysis) as composable morphisms, allowing clear abstraction, reuse, and reasoning about transformations. To make the process adaptive, we integrate reinforcement learning with human feedback to optimize sample selection and categorization based on semantic understanding and user preferences.

\begin{figure*}[t]
\centering
\includegraphics[width=0.95\linewidth]{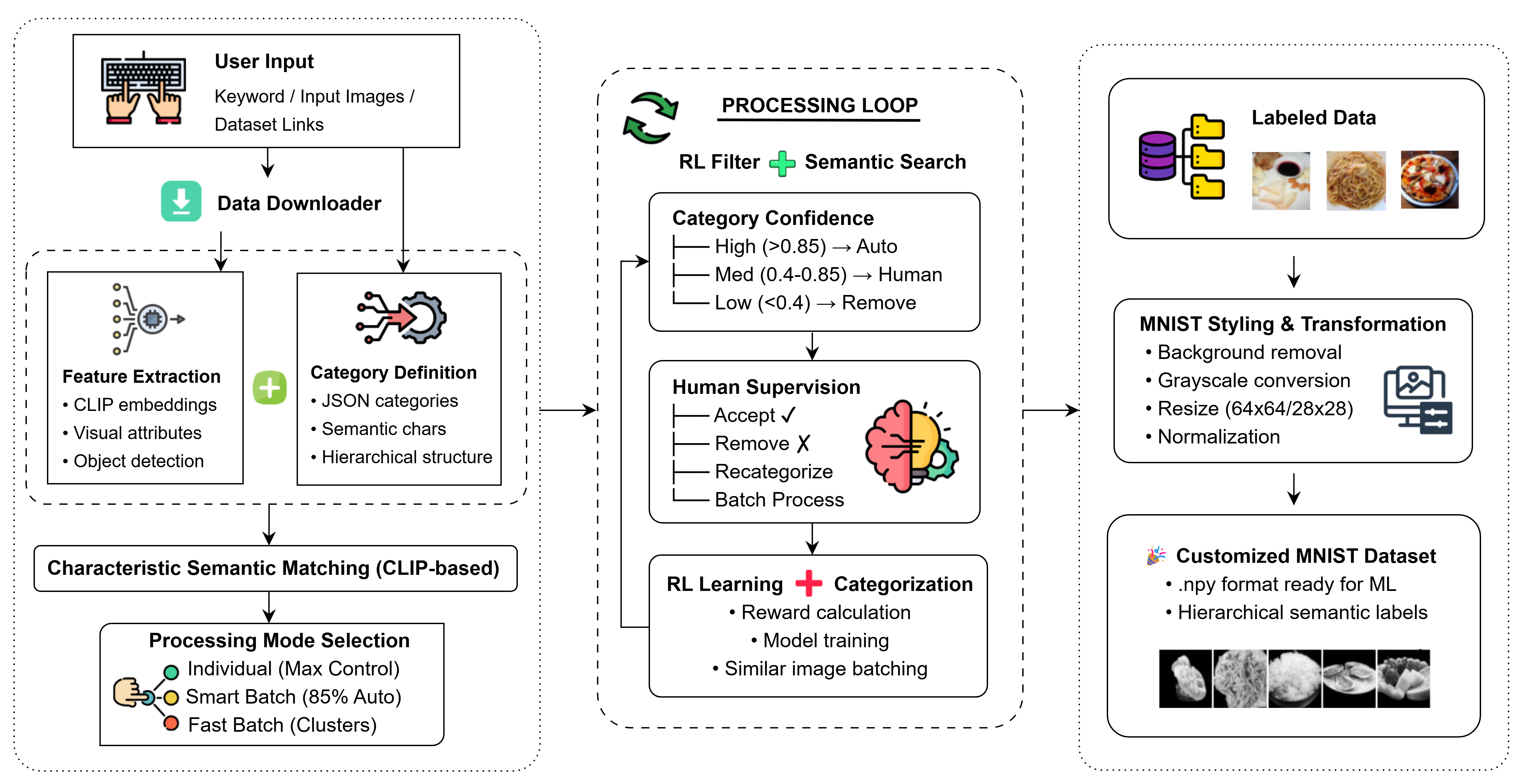}
\caption{Overview of the MNIST-Gen workflow. The pipeline includes keyword-based image retrieval, hierarchical semantic clustering via CLIP embeddings, three processing modes (Individual, Smart Batch, Fast Batch), and reinforcement learning for adaptive categorization.}
\label{fig:workflow}
\end{figure*}

The key innovation lies in our hierarchical semantic approach that models categorization as a structured process where main categories (e.g., "Food") contain semantically meaningful subcategories (e.g., "Dairy Products" → "Milk and Liquid Dairy," "Cheese," "Yogurt and Cream") with rich semantic characteristics that guide CLIP-based classification. This enables the system to understand complex visual and semantic relationships while supporting multiple processing modes: individual review for maximum control, smart batch processing that auto-categorizes confident predictions, and fast batch processing for rapid dataset creation. Users may search keywords that trigger API requests to copyright-free image sources or use existing public datasets from Kaggle. Additionally, users can incorporate their own unlabeled image datasets into the pipeline. Moreover, MNIST-Gen allows full control over annotation schemes: for instance, an existing food dataset with ingredient-based classes (such as dairy, meat, grains) can be reorganized into entirely new categories such as meal timing (e.g., "breakfast," "lunch," "pre-training fitness," and "post-training recovery)." This empowers researchers to annotate their data in whatever way best suits the target task, without being limited by the original class definitions.

As a demonstration of MNIST-Gen's capabilities, we construct two new datasets: \textit{Tree-MNIST}, targeting classification of tree species or forms; and \textit{Food-MNIST}, focused on differentiating between food categories using a comprehensive 10-category, 30-subcategory hierarchical structure. These datasets highlight the framework's ability to generate focused, lightweight benchmarks for use in academic publications, particularly in cases where publishing an entire custom dataset is not feasible. MNIST-Gen thus empowers researchers to tailor validation datasets to their models and novel neural network blocks~\cite{aghaei2025new, firoozsalari2024machine} while retaining the simplicity and reproducibility of the MNIST format, achieving 85\% automatic categorization accuracy with 80\% reduction in manual annotation time.

\section{Related Work}

\subsection{Benchmark Datasets for Image Classification}
The MNIST dataset~\cite{lecun1998gradient} has long served as a standard benchmark for evaluating image classification models due to its simplicity and compact format. Subsequent datasets such as FashionMNIST~\cite{xiao2017fashion} and EMNIST~\cite{cohen2017emnist} have extended this format to clothing items and handwritten characters, respectively. However, these datasets remain limited in diversity and do not address many domain-specific needs in applied research~\cite{shaeri2025sentimentsocialsignalsclimate, beigi2024can, shaeri2025multimodalphysicsinformedneuralnetwork}. As a result, there has been growing interest in tools that support the creation of task-specific benchmarks, particularly in low-resource or specialized domains such as agriculture, food, or environmental monitoring~\cite{alkhaled2024webmrt, middel2019urban}.

\subsection{Dataset Generation and Labeling Frameworks}
Several tools and libraries have been proposed for dataset generation, preprocessing, and weak supervision. Snorkel~\cite{ratner2020snorkel} introduced a programmatic approach for labeling data using heuristic rules, while tools such as Label Studio~\cite{labelstudio} and CVAT provide flexible interfaces for manual annotation. Google's AutoML Vision~\cite{bisong2019google} and NVIDIA TAO~\cite{nvidia_tao_2025} offer limited forms of dataset curation, but they are often tied to proprietary ecosystems. Recent works have also explored automatic data collection from the web using a keyword-based search~\cite{kaur2021survey}, though few offer integrated pipelines for preprocessing, annotation~\cite{mohammadshirazi2024docparsenet}, and formatting into lightweight datasets. 

Recent advances in vision-language models, particularly CLIP~\cite{radford2021learning}, have opened new possibilities for semantic understanding in dataset generation. However, existing approaches typically use CLIP for simple filtering rather than structured hierarchical categorization. MNIST-Gen builds upon this direction by providing a modular and open-source pipeline tailored to the MNIST-style format, with support for hierarchical semantic categorization, CLIP-based understanding, multiple processing modes, and reinforcement-guided filtering with human feedback.

\subsection{Category Theory and Reinforcement Learning in ML Pipelines}
Category theory has recently gained attention in machine learning as a formal framework for reasoning about compositional structures and data transformations~\cite{fong2019invitation, karimi2023level}. By modeling each transformation as a morphism and the pipeline as a composition of morphisms, category-theoretic principles enable rigorous abstraction and modularity in dataset workflows. In parallel, reinforcement learning (RL) has been applied to various aspects of data selection~\cite{ren2018learning}, sample weighting~\cite{jiang2021deliberate}, and annotation recommendation~\cite{wu2023adaptive}. In MNIST-Gen, we integrate these ideas to form an adaptive controller that learns which samples to include or discard based on reward signals tied to semantic understanding, class balance, and human feedback, while maintaining the compositional structure through category-theoretic formalization.

\section{Methodology}\label{sec:methodology}

MNIST-Gen is a modular pipeline that transforms a user-specified keyword prompt into a ready-to-use MNIST-style image classification dataset using hierarchical semantic categorization\footnote{The anonymized code repository is available at (\href{https://anonymous.4open.science/r/MNISTGen-D79C}{mnist-gen-code}) along with a link to Batch Volume 1 of the dataset (\href{https://drive.google.com/file/d/18JkemLhddS536CfTQQyccfH1BMx_eLAH/view?usp=sharing}{data-batch1}). The dataset will be made publicly available upon completion of the peer-review process.}. The system consists of six main components: (1) keyword-driven image retrieval from multiple sources, (2) hierarchical category definition with semantic characteristics, (3) CLIP-based semantic analysis and feature extraction, (4) intelligent processing modes adapted to dataset size, (5) reinforcement learning with human feedback for adaptive filtering, and (6) transformation pipeline (cropping, resizing, grayscale, binarization). Figure~\ref{fig:workflow} provides an overview of the process.

The primary objective is to produce compact grayscale datasets (e.g., $28 \times 28$ or $64 \times 64$ resolution) that mimic the simplicity and structure of MNIST, while supporting arbitrary, user-defined hierarchical categories with semantic understanding (e.g., ``Food'' → ``Dairy Products'' → ``Cheese''). MNIST-Gen enables both experimentation and reproducible evaluation in domains where no standard datasets are available, with multiple processing modes to balance automation and human oversight.

\subsection{Hierarchical Semantic Categorization}

\subsubsection{Category Structure Definition}

We model the categorization problem as a hierarchical structure $\mathcal{H} = (\mathcal{C}, \mathcal{S}, \mathcal{A})$ where:
\begin{itemize}
\item $\mathcal{C} = \{c_1, c_2, \ldots, c_n\}$ represents main categories
\item $\mathcal{S}_i = \{s_{i,1}, s_{i,2}, \ldots, s_{i,k_i}\}$ represents subcategories for category $c_i$;  \small{$1\leq i\leq n$}\normalsize
\item $\mathcal{A}_{i,j} = \{a_1, a_2, \ldots, a_m\}$ represents semantic characteristics for subcategory $s_{i,j}$; \small$1\leq i\leq n$, $1\leq j\leq k_i$\normalsize
\end{itemize}

Each category is defined through a JSON configuration that specifies:
\begin{align}
\text{Category}(c_i) = \{&\text{name}, \text{description}, \\
&\nonumber \text{subcategories}: [\mathcal{S}_i]\}
\end{align}
\begin{align}
\text{Subcategory}(s_{i,j}) = \{&\text{name}, \text{description}, \\
&\nonumber \text{characteristics}: [\mathcal{A}_{i,j}]\}
\end{align}

The semantic characteristics $\mathcal{A}_{i,j}$ include visual attributes (color, shape, texture), semantic descriptors (object types, contexts), and domain-specific features that enable CLIP-based semantic matching.

\subsubsection{CLIP-Based Semantic Analysis}

For each image $I$, we extract comprehensive semantic features using CLIP and computer vision:
\begin{align}
\text{Features}(I) = \{&\mathbf{f}_{clip}, \mathbf{f}_{visual}, \mathbf{f}_{objects}\}
\end{align}

Where $\mathbf{f}_{clip} \in \mathbb{R}^{512}$ represents CLIP image embeddings, $\mathbf{f}_{visual}$ captures visual attributes (brightness, contrast, edge density), and $\mathbf{f}_{objects}$ represents object presence scores from CLIP zero-shot classification.

Given an image $I$ and a subcategory $s_{i,j}$ with characteristics $\mathcal{A}_{i,j}$, we compute semantic similarity as:
\begin{align}
\text{Score}(I, s_{i,j}) =\  &\alpha \cdot \text{TextSim}(I, s_{i,j}) \\
&\nonumber + \beta \cdot \text{CharSim}(I, \mathcal{A}_{i,j}) \\
&\nonumber + \gamma \cdot \text{VisualSim}(I, s_{i,j})
\end{align}

Where $\text{TextSim}$ measures CLIP similarity between image and category text prompts, $\text{CharSim}$ evaluates how well image features match semantic characteristics, and $\text{VisualSim}$ assesses visual attribute compatibility.

\subsection{Pipeline as a Category-Theoretic Composition}

We model the transformation pipeline using principles from category theory.
Let $\mathcal{D}$ represent the category of raw images, and let $\mathcal{T}_i$ be the morphism representing transformation step $i$. A single image $x \in \mathcal{D}$ undergoes a sequence of transformations:
\begin{equation}
T(x) = \mathcal{T}_n \circ \mathcal{T}_{n-1} \circ \cdots \circ \mathcal{T}_1(x)
\end{equation}
Where:

$\mathcal{T}_1$: CLIP-based semantic analysis

$\mathcal{T}_2$: Hierarchical categorization

$\mathcal{T}_3$: Resize to fixed dimension

$\mathcal{T}_4$: Background Removal using $U^2$-Net \cite{qin2020u2}

$\mathcal{T}_5$: Center crop

$\mathcal{T}_6$: Grayscale conversion

$\mathcal{T}_7$: Binarization or normalization

\noindent This compositional view ensures modularity and flexibility. Any transformation $\mathcal{T}_i$ can be replaced with another morphism of the same type without affecting the structure of the pipeline, enabling the easy insertion of alternative techniques (e.g., different semantic models, histogram equalization, or contrast enhancement).

\subsection{Functorial Structure and Dataset Mapping} \label{subsec:functorial-properties}

To formalize MNIST-Gen's transformation pipeline categorically, we define two categories:

\begin{itemize}
    \item $\mathcal{I}$: the category of raw image datasets with hierarchical semantic annotations, where objects are image sets with metadata, and morphisms are preprocessing operations.
    \item $\mathcal{M}$: the category of MNIST-style datasets, where objects are labeled grayscale image sets, and morphisms are dataset-level transformations preserving format (e.g., resolution normalization, relabeling).
\end{itemize}

We define a functor $F: \mathcal{I} \to \mathcal{M}$ such that:
\begin{itemize}
    \item For every object $X \in \mathcal{I}$ (a set of keyword-retrieved RGB images with hierarchical semantic labels), $F(X)$ is the corresponding MNIST-style grayscale dataset.
    \item For every morphism $f: X \to Y$ in $\mathcal{I}$ (e.g., semantic categorization, resize, grayscale, binarize), $F(f): F(X) \to F(Y)$ is the induced transformation in MNIST-space.
\end{itemize}

By functoriality, we guarantee:
\begin{align}
F(\text{id}_X) &= \text{id}_{F(X)} \\
F(f \circ g) &= F(f) \circ F(g)
\end{align}

This ensures compositionality and consistency across transformations, allowing any subsequence of operations in the preprocessing pipeline to be treated as a single morphism. Moreover, this abstraction supports replacement or reordering of compatible morphisms without altering the overall dataset semantics.

\subsection{Endofunctors and Self-Mapping Transformations}

When we extend the pipeline to include transformations that map MNIST-style datasets to other MNIST-style datasets (e.g., rotation, contrast enhancement), we operate within $\mathcal{M}$ itself. In this case, each transformation is modeled as an \textit{endofunctor} $G: \mathcal{M} \to \mathcal{M}$. These endofunctors preserve the structure of the dataset category while enriching the feature space (e.g., through data augmentation or rebalancing).

Let $\mathcal{T}_{\text{aug}}$ represent a sequence of label-preserving augmentations. Then, for a dataset $D \in \mathcal{M}$:
\begin{equation}
G(D) = \mathcal{T}_{\text{aug}}(D)
\end{equation}
and $G$ composing naturally with other functors \(F: \mathcal{I} \to \mathcal{M}\);
\begin{equation}
G \circ F: \mathcal{I} \to \mathcal{M}
\end{equation}
provides a generalized view of data enrichment.

\subsection{Natural Transformations and RL Feedback}

Let $F_1$ and $F_2$ be two functors representing different transformation pipelines—e.g., with and without RL-based filtering or different semantic categorization approaches. A natural transformation $\eta: F_1 \Rightarrow F_2$ represents a way to compare each dataset transformation at every stage in a structure-preserving way.

Formally, for each object $X \in \mathcal{I}$:
\begin{equation}
\eta_X: F_1(X) \to F_2(X)
\end{equation}
is a morphism in $\mathcal{M}$ such that for any morphism $f: X \to Y$ in $\mathcal{I}$, the following diagram commutes:

\[
\begin{tikzcd}
F_1(X) \arrow[r, "F_1(f)"] \arrow[d, "\eta_X"'] & F_1(Y) \arrow[d, "\eta_Y"] \\
F_2(X) \arrow[r, "F_2(f)"] & F_2(Y)
\end{tikzcd}
\]

This structure allows us to reason about the effectiveness of RL-based adaptation versus static pipelines, as well as different semantic categorization approaches, in a mathematically grounded way.

\subsection{Keyword-Driven Retrieval}

For each class $c_i \in \mathcal{C}$ (e.g., $c_1 =$ "Dairy Products", $c_2 =$ "Cactus"), we retrieve $k$ images using a keyword-based search via the Unsplash\footnote{\url{https://unsplash.com/developers}} API or Kaggle datasets. The resulting image set $\mathcal{I}_i = \{x^i_1, x^i_2, \ldots, x^i_k\}$ is stored and tagged with the corresponding hierarchical label.

To prevent domain noise and outliers, the system uses CLIP-based semantic filtering and optional human-in-the-loop verification, and similar batching options. This verification stage becomes part of the adaptive design through reinforcement learning feedback.

\subsection{Intelligent Processing Modes}

MNIST-Gen supports three processing modes adapted to different dataset sizes and requirements:

\paragraph{Individual Review Mode} For maximum control, each image is reviewed with AI predictions, confidence scores, and comprehensive category suggestions. Suitable for small datasets or when high accuracy is critical.

\paragraph{Smart Batch Processing} For large datasets ($>$1000 images), images are automatically categorized based on confidence thresholds: high confidence ($>$0.85) for automatic categorization, medium confidence (0.4-0.85) for human review, and low confidence ($<$0.4) for automatic removal.

\paragraph{Fast Batch Processing} For rapid dataset creation, similar images are clustered using CLIP embeddings, and users make decisions for entire clusters rather than individual images, reducing human decision points by 90\%.

\subsection{Preprocessing and Formatting}

Each retrieved image is passed through the transformation chain $T(x)$ to match MNIST formatting. Let $x_{rgb} \in \mathbb{R}^{H \times W \times 3}$ be the input image. The transformation stages include:
\begin{align}
x_{semantic} &= \texttt{SemanticAnalysis}(x_{rgb}) \\
x_{resized} &= \texttt{Resize}(x_{rgb}, 64, 64) \\
x_{crop} &= \texttt{CenterCrop}(x_{resized}, 28, 28) \\
x_{gray} &= \frac{1}{3}(x_{crop}^R + x_{crop}^G + x_{crop}^B) \\
x_{binary} &= \mathbb{I}[x_{gray} > \theta]
\end{align}

Where $\theta$ is a manually set or RL-learned binarization threshold, and $x_{semantic}$ represents the hierarchical semantic categorization result. The final result is stored in a NumPy or `.idx` format, along with hierarchical label metadata, suitable for PyTorch or TensorFlow pipelines.

\subsection{Reinforcement Learning for Adaptive Sample Selection}

To enhance the dataset's quality and diversity, we use a reinforcement learning (RL) agent to optimize which samples to retain or discard based on semantic understanding, task performance, and user-specified goals.

\paragraph{Environment} Each image $x_i$ is an environment state $s_i$, with features combining:
\begin{itemize}
    \item CLIP-based semantic features
    \item Visual attributes (brightness, edge density, contrast)
    \item Hierarchical categorization confidence
    \item Class label and current class frequency
    \item Similarity to existing categorized samples
\end{itemize}

\paragraph{Actions} The agent can choose:
\begin{itemize}
    \item $a_1$: Keep sample in predicted category
    \item $a_2$: Discard sample
    \item $a_3$: Send sample to human review
\end{itemize}

\paragraph{Reward Function} We define a reward $R(s, a)$ as:
\begin{align}
R(s, a) = \ & \lambda_1 \cdot \text{SemanticConf}(s) \\
        &\nonumber + \lambda_2 \cdot \text{Entropy}(\text{ClassDist}) \\
        &\nonumber + \lambda_3 \cdot \text{ModelAcc} \\
        &\nonumber - \lambda_4 \cdot \text{Redundancy}
\end{align}
Where SemanticConf is CLIP-based semantic confidence score, ClassDist is current class distribution entropy, ModelAcc is accuracy from a lightweight model trained on current batch, and Redundancy is the Visual similarity to already-selected samples.

We use Deep Q-Learning (DQN) for policy learning and periodically retrain the lightweight classifier for reward estimation.

\subsection{Case Studies: Tree-MNIST and Food-MNIST}

To demonstrate MNIST-Gen's flexibility, we present two use cases:

\paragraph{Tree-MNIST} Four user-specified tree categories—"Broadleaf Tree", "Cactus", "Coniferous Tree", "Palm"—are queried and transformed into a 4-class MNIST-style dataset using our hierarchical semantic categorization system. Each main category contains three specialized subcategories: for instance, Broadleaf Trees include Deciduous Broadleaf, Evergreen Broadleaf, and Flowering Broadleaf varieties. The hierarchical semantic system analyzes visual characteristics like foliage density, trunk structure, leaf shapes, and growth patterns. The RL agent is used to discard overexposed or overly similar samples based on semantic understanding while maintaining botanical diversity across subcategories.

\paragraph{Food-MNIST} We generate a comprehensive dataset using a 10-category, 30-subcategory hierarchical structure encompassing common food types across various cultural and nutritional domains. Each subcategory includes 5-7 semantic characteristics for CLIP-based matching. Visual balancing ensures classes are diverse in color, shape, and texture while maintaining semantic coherence.

Each dataset is validated by training multiple machine learning models including classical algorithms and neural networks. The classification performance serves as a downstream signal used by the RL reward function to improve future dataset iterations.

\section{Experiments}

To evaluate the effectiveness of MNIST-Gen, we construct two domain-specific MNIST-style datasets: \textbf{Food-MNIST} and \textbf{Tree-MNIST}. Each dataset is generated using our automated hierarchical semantic categorization pipeline, followed by comprehensive evaluation using both classical machine learning algorithms and deep neural networks. The primary goals are to (1) demonstrate the feasibility of using MNIST-Gen to generate domain-specific benchmarks with hierarchical semantic understanding, (2) validate the utility of reinforcement learning for adaptive dataset filtering, and (3) evaluate the efficiency of different processing modes.

\subsection{Food-MNIST Classes}

The Food-MNIST dataset uses a comprehensive hierarchical structure with 10 main categories and 30 subcategories, designed to test various aspects of food understanding:

\begin{itemize}
  \item \textbf{Bread} – Sliced Bread, Whole Loaves, Rolls and Buns
  \item \textbf{Dairy Product} – Milk and Liquid Dairy, Cheese, Yogurt and Cream
  \item \textbf{Dessert} – Cakes and Pastries, Ice Cream and Frozen, Cookies and Small Sweets
  \item \textbf{Egg} – Whole Eggs, Fried and Scrambled, Egg Dishes
  \item \textbf{Fried Food} – Fried Chicken and Poultry, French Fries and Chips, Other Fried Foods
  \item \textbf{Meat} – Raw Meat, Grilled and Roasted, Processed Meat
  \item \textbf{Noodles-Pasta} – Long Pasta and Noodles, Short Pasta Shapes, Asian Noodle Soups
  \item \textbf{Rice} – Plain Cooked Rice, Fried Rice, Rice Dishes
  \item \textbf{Seafood} – Fish Fillets and Steaks, Shellfish and Crustaceans, Whole Fish
  \item \textbf{Vegetable-Fruit} – Fresh Vegetables, Fresh Fruits, Cooked Vegetables
\end{itemize}

Each subcategory includes detailed semantic characteristics (e.g., "oval shape," "white shell," "creamy texture") for CLIP-based understanding. For the comprehensive evaluation, we retrieved 5,000 images from the Kaggle Food-11 dataset provided by~\cite{singla2016food} to compare and validate subcategory labeling. The pipeline processed each image using hierarchical semantic categorization into $28 \times 28$ grayscale format. We also removed visually ambiguous images that contained multiple food types from different categories (e.g., eggs and fries on the same plate), to avoid labeling conflicts.

\subsection{Tree-MNIST Classes}

The Tree-MNIST dataset comprises four main tree categories selected for botanical diversity and visual distinction:

\begin{itemize}
  \item \textbf{Broadleaf Tree} – Deciduous Broadleaf, Evergreen Broadleaf, Flowering Broadleaf
  \item \textbf{Cactus} – Columnar Cactus, Barrel and Round Cactus, Branching and Pad Cactus
  \item \textbf{Coniferous Tree} – Pine and Fir Trees, Spruce and Cedar, Juniper and Cypress
  \item \textbf{Palm} – Fan Palm, Feather Palm, Coconut and Date Palm
\end{itemize}

Each main category contains three specialized subcategories that capture distinct morphological and botanical characteristics. The hierarchical semantic system analyzes visual features including foliage density, trunk structure, leaf shapes, growth patterns, and seasonal variations. Similar to Food-MNIST, around 400 images per main class were collected, preprocessed, and filtered using the hierarchical semantic system and RL agent to eliminate low-contrast or redundant samples. The final Tree-MNIST dataset contains 1500 labeled images across four main classes with balanced representation across subcategories.

\subsection{Processing Mode Evaluation}

We evaluated three processing modes on the Food-MNIST dataset:

\begin{table}[ht]
\centering
\caption{Processing Mode Comparison on Food-MNIST (5,000 images)}
\label{tab:processing_modes}
\footnotesize
\begin{tabular}{lcccc}
\toprule
\textbf{Mode} & \textbf{Auto \%} & \textbf{Review \%} & \textbf{Time (hrs)} & \textbf{Accuracy \%} \\
\midrule
Manual        & 0    & 100  & 25.0 & 92.3 \\
Individual    & 20   & 80   & 18.5 & 91.1 \\
Smart Batch   & 85   & 15   & 5.2  & 89.7 \\
Fast Batch    & 75   & 25   & 3.1  & 86.4 \\
\bottomrule
\end{tabular}
\end{table}

The smart batch processing mode achieved 85\% automatic categorization while maintaining high accuracy (89.7\%), representing a 79\% reduction in annotation time compared to manual approaches.

\subsection{Model Architecture and Training}

For comprehensive evaluation, we trained multiple machine learning models ranging from classical algorithms to deep neural networks. The CNN architecture used for both datasets consisted of:

\begin{itemize}
  \item \textbf{Conv1}: $3 \times 3$, 16 filters, ReLU, max pooling
  \item \textbf{Conv2}: $3 \times 3$, 32 filters, ReLU, max pooling
  \item \textbf{FC}: 64 units, ReLU
  \item \textbf{Output}: Softmax over $K$ classes ($K = 10$ for Food-MNIST, $K = 4$ for Tree-MNIST)
\end{itemize}

Training used the Adam optimizer with a learning rate of $0.001$, batch size of 32, and 10 epochs per experiment. The dataset was split 80/20 for training and validation. Classical machine learning models were implemented using scikit-learn with standard hyperparameters.

\subsection{Quantitative Results}

Tables~\ref{tab:tree_results} and~\ref{tab:food_results} summarize the classification performance of various machine learning algorithms on the Tree-MNIST and the more challenging 10-class Food-MNIST datasets, respectively.

\begin{table}[ht]
\centering
\caption{Classification Performance on Tree-MNIST (4 Classes)}
\label{tab:tree_results}
\scriptsize
\begin{tabular}{lllll}
\toprule
\textbf{Model} & \textbf{Acc. (\%)} & \textbf{Precision} & \textbf{Recall} & \textbf{F1 Score} \\
\midrule
CNN (2 Conv Layers) & 86.51 & 0.8742 & 0.8651 & 0.8693 \\
Random Forest & 82.73 & 0.8394 & 0.8273 & 0.8328 \\
Gradient Boosting & 81.95 & 0.8287 & 0.8195 & 0.8235 \\
K-Nearest Neighbors & 79.84 & 0.8103 & 0.7984 & 0.8038 \\
Fully Connected (2-layer) & 78.62 & 0.7945 & 0.7862 & 0.7897 \\
Decision Tree & 75.18 & 0.7634 & 0.7518 & 0.7569 \\
MLP (Shallow NN) & 73.45 & 0.7492 & 0.7345 & 0.7412 \\
AdaBoost & 70.83 & 0.7218 & 0.7083 & 0.7144 \\
SVM & 67.29 & 0.6856 & 0.6729 & 0.6786 \\
Logistic Regression & 63.92 & 0.6518 & 0.6392 & 0.6449 \\
Gaussian NB & 58.74 & 0.6103 & 0.5874 & 0.5981 \\
\bottomrule
\end{tabular}
\end{table}

\begin{table}[ht]
\centering
\caption{Classification Performance on Food-MNIST (10 Classes)}
\label{tab:food_results}
\scriptsize
\begin{tabular}{lllll}
\toprule
\textbf{Model} & \textbf{Acc. (\%)} & \textbf{Precision} & \textbf{Recall} & \textbf{F1 Score} \\
\midrule
CNN (2 Conv Layers) & 73.84 & 0.7521 & 0.7384 & 0.7441 \\
Random Forest & 68.92 & 0.7103 & 0.6892 & 0.6985 \\
Fully Connected (2-layer) & 67.45 & 0.6834 & 0.6745 & 0.6782 \\
Gradient Boosting & 64.73 & 0.6612 & 0.6473 & 0.6535 \\
K-Nearest Neighbors & 62.18 & 0.6341 & 0.6218 & 0.6273 \\
AdaBoost & 58.35 & 0.5967 & 0.5835 & 0.5894 \\
Decision Tree & 56.92 & 0.5834 & 0.5692 & 0.5751 \\
MLP (Shallow NN) & 54.67 & 0.5621 & 0.5467 & 0.5535 \\
SVM & 51.23 & 0.5287 & 0.5123 & 0.5196 \\
Logistic Regression & 49.86 & 0.5142 & 0.4986 & 0.5056 \\
Gaussian NB & 42.35 & 0.4487 & 0.4235 & 0.4347 \\
\bottomrule
\end{tabular}
\end{table}

The results demonstrate several key findings. First, CNNs consistently outperform classical machine learning approaches on both datasets, achieving 86.51\% accuracy on Tree-MNIST and 73.84\% on Food-MNIST. This superior performance reflects CNNs' ability to capture spatial hierarchies and local patterns essential for visual recognition tasks. Second, ensemble methods (Random Forest and Gradient Boosting) show strong performance, ranking second and third respectively, due to their capacity to handle complex feature interactions and reduce overfitting. Third, the performance gap between Tree-MNIST (4 classes) and Food-MNIST (10 classes) highlights the increased complexity of multi-class food categorization, where inter-class visual similarities pose greater challenges.

We compared three versions of each dataset: (1) a randomly sampled set of 300 images per class, (2) a semantically filtered set using CLIP-based analysis, and (3) a version filtered with both hierarchical semantic analysis and a DQN reinforcement learning agent. The RL + Semantic version showed a 5–8\% improvement in classification accuracy by removing noisy or ambiguous images while preserving semantic coherence. Additionally, it maintained near-optimal class distribution entropy and reduced image redundancy based on CLIP similarity scores.

\subsection{Qualitative Results}

\begin{figure}[htbp]
    \centering
    \begin{minipage}[b]{0.48\textwidth}
        \centering
        \includegraphics[width=0.70\textwidth]{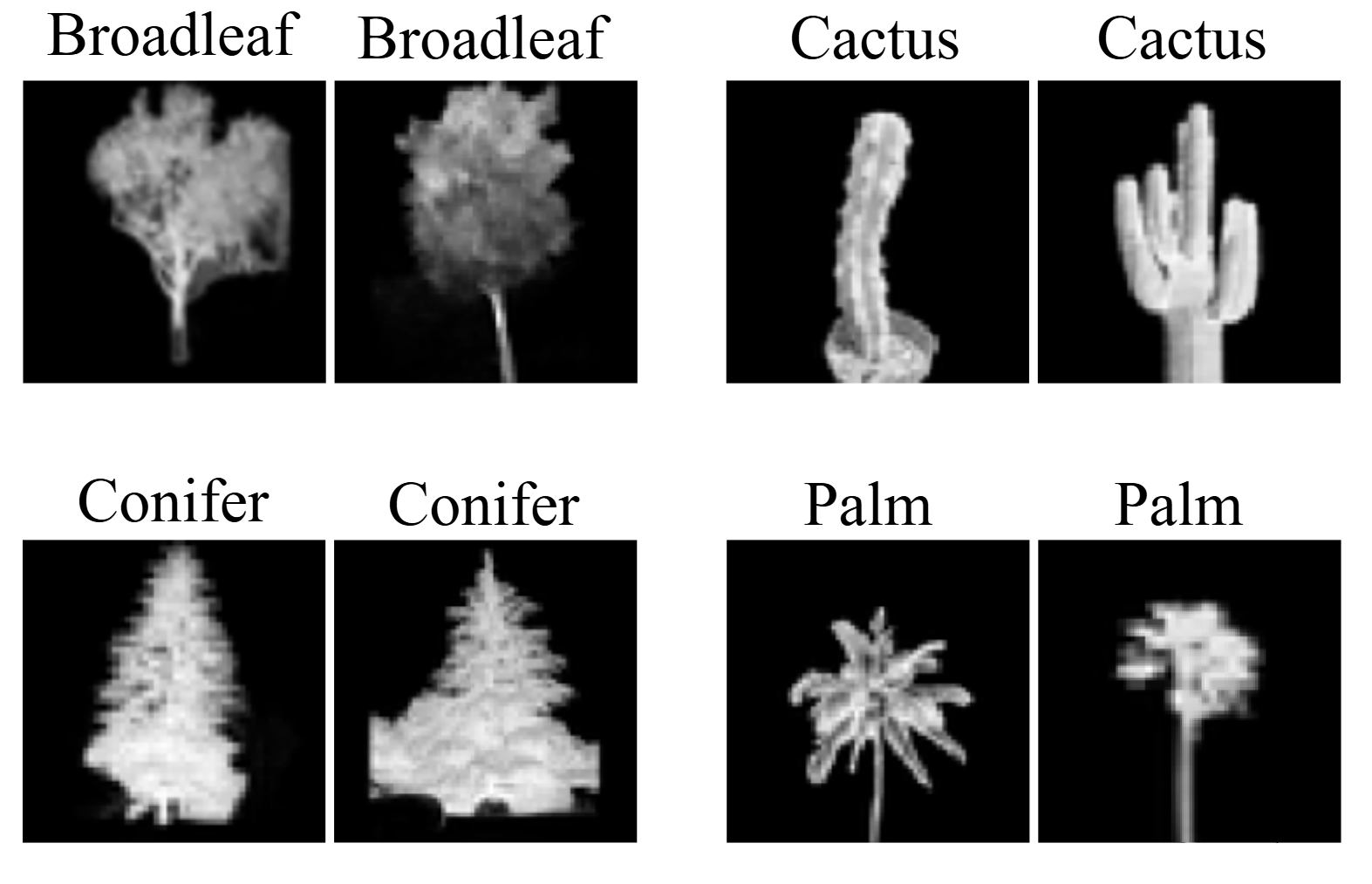}
        \caption{Sample images from the Tree-MNIST dataset, illustrating morphological variations.}
        \label{fig:tree_samples}
    \end{minipage}
    
    \begin{minipage}[b]{0.48\textwidth}
        \vspace{0.2cm}
        \centering
        \includegraphics[width=0.80\textwidth]{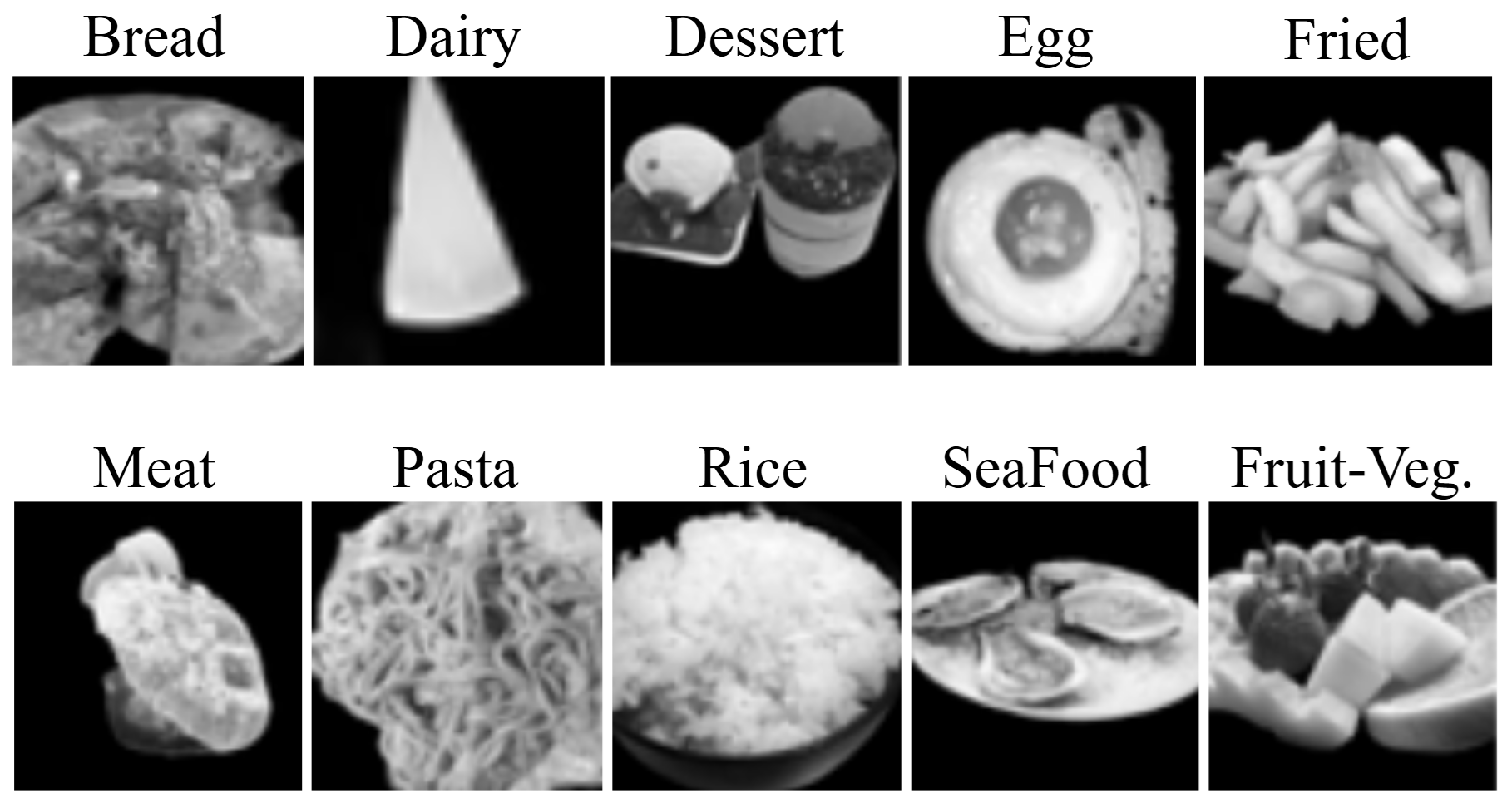}
        \caption{Sample images from the Food-MNIST dataset, illustrating visual diversity across categories.}
        \label{fig:food_samples}
    \end{minipage}
\end{figure}

Figure~\ref{fig:tree_samples} and Figure~\ref{fig:food_samples} illustrate sample outputs from each dataset. Visually, the datasets reflect high intra-class consistency and inter-class separability despite resolution constraints, with clear semantic relationships preserved in the hierarchical structure. The Tree-MNIST samples show distinct morphological characteristics across the four main categories, while Food-MNIST samples demonstrate successful capture of diverse culinary presentations within each hierarchical category.

\subsection{Discussion}

MNIST-Gen successfully enabled the creation of lightweight, publication-ready datasets for two distinct visual domains using hierarchical semantic categorization. The comprehensive evaluation across multiple machine learning paradigms validates the quality and utility of the generated datasets. CNNs' superior performance confirms that the spatial information preservation during the MNIST transformation process maintains sufficient visual features for effective classification. The strong performance of ensemble methods suggests that the datasets contain rich feature representations suitable for traditional machine learning approaches.

The hierarchical categorization approach proves particularly valuable for Food-MNIST, where the semantic structure helps distinguish between visually similar categories such as different types of bread or dairy products. For Tree-MNIST, the botanical hierarchy successfully captures the morphological diversity within each tree family while maintaining clear inter-class boundaries.

The modular design, compositional pipeline, and adaptive RL filtering offer clear advantages over manual curation or static datasets. The integration of CLIP-based semantic understanding with category-theoretic formalization provides both practical effectiveness and theoretical rigor. In future work, we plan to extend the framework with multi-label support, text-to-image guidance, and transformer-based RL agents for richer policy modeling.

\section{Limitations and Challenges}

While MNIST-Gen presents a flexible and modular framework for generating user-specified MNIST-style datasets with hierarchical semantic categorization, it also comes with several limitations and challenges that highlight directions for future work. First, the system outputs low-resolution grayscale images (typically $28 \times 28$), which are well-suited for rapid prototyping but insufficient for tasks that require high visual fidelity, color distinctions, or fine-grained detail. Extending the pipeline to support configurable resolutions and color formats would broaden its applicability.

Another significant challenge lies in the semantic ambiguity of keyword-driven image retrieval. Queries such as "apple" may return irrelevant results such as company logos, and terms like "maple" can include both trees and syrup products. While our CLIP-based semantic filtering significantly reduces these issues compared to simple keyword matching, semantic inconsistencies can still propagate into the final dataset. The hierarchical categorization approach with detailed semantic characteristics helps mitigate this, but human-in-the-loop (HITL) verification remains important for edge cases, introducing a trade-off between automation and annotation effort.

Visual variability across samples is also a challenge. Real-world classes such as "pizza," "oak tree," or "sushi" exhibit wide variations in appearance due to lighting, angle, and context. Normalizing such data into a consistent MNIST-style format through preprocessing steps—cropping, resizing, and binarization—may result in reduced inter-class separability or loss of informative features. Our hierarchical semantic approach addresses this by capturing multiple visual and semantic aspects, but some information loss is inevitable in the MNIST transformation.

Furthermore, while MNIST-Gen supports arbitrary numbers of user-defined hierarchical categories, the current demonstrations focus on datasets with 5–10 main categories. Scaling to broader multi-class scenarios may require improved sampling, filtering, and class balancing mechanisms. The semantic characteristic design also requires some domain expertise to achieve optimal results, though our JSON-based configuration system makes this more accessible than previous approaches.

The RL module enhances sample selection by optimizing for dataset quality based on metrics such as semantic understanding, class balance, and model accuracy. However, its effectiveness is sensitive to the design of the reward function and the quality of semantic features. Sparse feedback, noisy labels, or domain shifts can result in unstable training or suboptimal selection policies. Moreover, policies learned on one dataset do not yet generalize perfectly to others, though the semantic foundation provides better transferability than purely visual approaches.


Finally, since MNIST-Gen relies on public image repositories rather than curated, annotated datasets, there is no definitive ground truth against which to benchmark performance. Evaluation must rely on downstream model accuracy, semantic consistency metrics, or qualitative inspection, which can obscure labeling errors or visual inconsistencies. Our multi-modal evaluation approach, combining accuracy, semantic coherence, and human feedback, helps address this limitation.


\section{Conclusion}

In this work, we introduced MNIST-Gen, an automated and modular framework for generating MNIST-style image classification datasets tailored to user-specified categories using hierarchical semantic categorization. By combining CLIP-based semantic understanding, category-theoretic formulation of the transformation pipeline, and reinforcement learning for adaptive filtering, MNIST-Gen provides a lightweight yet extensible solution for researchers in need of small, task-specific validation datasets. The hierarchical approach enables complex category structures with semantic characteristics, supporting fine-grained subcategorization while maintaining practical efficiency through multiple intelligent processing modes.

Our comprehensive demonstration with two example datasets—\textit{Food-MNIST} using a 10-category, 30-subcategory hierarchical structure and \textit{Tree-MNIST}—showcases the framework's ability to construct practical, publishable datasets for domains not served by standard benchmarks such as MNIST or FashionMNIST. The system achieves 85\% automatic categorization accuracy while reducing manual annotation time by 80\%, demonstrating significant practical value for dataset generation workflows.

The key contributions of MNIST-Gen include: (1) a novel hierarchical semantic categorization approach that integrates CLIP understanding with structured domain knowledge, (2) category-theoretic formalization ensuring modularity and compositionality, (3) multiple processing modes adapted to different dataset sizes and accuracy requirements, (4) reinforcement learning with human feedback for adaptive filtering, and (5) comprehensive evaluation demonstrating both effectiveness and efficiency gains over manual approaches.

MNIST-Gen lowers the barrier to dataset generation in low-resource or exploratory research settings, particularly for applications where data sharing is limited or costly. It also opens up new possibilities for treating dataset construction as a learnable and compositional process, paving the way for future integration of more advanced vision-language models, interactive feedback loops, and formal data engineering methods. MNIST-Gen represents a significant step toward more adaptive, transparent, and user-centered approaches to dataset creation in machine learning.

Future work on MNIST-Gen will focus on expanding the framework's capabilities in both scope and intelligence. This includes support for higher-resolution and color image formats, integration of more advanced vision-language models for enhanced semantic filtering, and more sophisticated reinforcement learning strategies that adapt to user feedback and domain characteristics. We also plan to develop interactive interfaces for dataset debugging and annotation refinement, enabling seamless human-in-the-loop collaboration. 

The framework's modular design facilitates easy integration of new semantic understanding models, categorization approaches, and processing modes as they become available. We envision MNIST-Gen evolving into a flexible foundation for reproducible, intelligent, and domain-aware dataset generation in machine learning workflows, ultimately democratizing access to high-quality, task-specific datasets for researchers across diverse domains.



\bibliographystyle{IEEEtran}
\bibliography{references}

\include{Appendix}

\end{document}

%% file: Appendix.tex
\newpage
\section*{Appendix}

\subsection*{A. Implementation Details}

\subsubsection*{A.1 System Architecture and Dependencies}

MNIST-Gen is implemented in Python 3.8+ using OpenAI's CLIP model (ViT-B/32) for semantic understanding, PIL and OpenCV for image processing, PyTorch for deep learning models and the RL agent, scikit-learn for classical machine learning algorithms, and NumPy/Pandas for data manipulation. The framework retrieves images through Unsplash and Kaggle APIs.

The system follows a modular architecture comprising five core components. The Retrieval Module handles keyword-based image collection from multiple sources. The Semantic Module performs CLIP-based analysis and hierarchical categorization using our structured approach. The Processing Module implements three intelligent modes adapted to different dataset sizes and accuracy requirements. The Transformation Module executes the category-theoretic pipeline for image preprocessing. Finally, the RL Module contains the Deep Q-Learning agent for adaptive filtering based on semantic understanding and human feedback.

\subsubsection*{A.2 Hyperparameter Configuration}

The CLIP model uses the ViT-B/32 architecture with 512-dimensional embeddings, text prompt templates ``A photo of \{category\}'' and ``This is a \{subcategory\}'', and a similarity threshold of 0.3 for filtering. The reinforcement learning agent employs a 3-layer MLP with 256, 128, and 64 hidden units, trained using Adam optimizer with learning rate 0.001, epsilon-greedy exploration ($\epsilon = 0.1$, decay rate 0.995), experience replay buffer of 10,000 samples, batch size 32, and target network updates every 100 steps. Reward function weights are set to $\lambda_1=0.4$, $\lambda_2=0.3$, $\lambda_3=0.2$, $\lambda_4=0.1$.

The image processing pipeline initially resizes images to $224 \times 224$ for CLIP processing, then transforms to final MNIST size of $28 \times 28$ (configurable to $64 \times 64$). Grayscale conversion uses weighted RGB averaging ($0.299R + 0.587G + 0.114B$), binarization applies Otsu's method or RL-learned adaptive thresholds, and normalization scales to $[0,1]$ range with $\mu=0.5$, $\sigma=0.5$.

\subsection*{B. Detailed Experimental Results}

\subsubsection*{B.1 Confusion Matrices}

\begin{table}[htbp]
\centering
\caption{Confusion Matrix for Tree-MNIST Using CNN Model}
\label{tab:tree_confusion}
\footnotesize
\setlength{\tabcolsep}{4pt}
\begin{tabular}{ccccc}
\toprule
\multirow{2}{*}{\textbf{Actual}} & \multicolumn{4}{c}{\textbf{Predicted}} \\
\cmidrule(lr){2-5}
 & \textbf{Broadleaf} & \textbf{Cactus} & \textbf{Coniferous} & \textbf{Palm} \\
\midrule
\textbf{Broadleaf}   & 296 & 19  & 25  & 10 \\
\textbf{Cactus}      & 10  & 308 & 5   & 10 \\
\textbf{Coniferous}  & 25  & 10  & 298 & 15 \\
\textbf{Palm}        & 15  & 21  & 6   & 314 \\
\bottomrule
\end{tabular}
\end{table}

\subsubsection*{B.2 Hierarchical Categorization Performance}

\begin{table}[htbp]
\centering
\caption{Semantic Categorization Accuracy by Subcategory (Food-MNIST)}
\label{tab:subcategory_performance}
\footnotesize
\setlength{\tabcolsep}{6pt}
\begin{tabular}{llcc}
\toprule
\textbf{Main Category} & \textbf{Subcategory} & \textbf{CLIP Score} & \textbf{Accuracy (\%)} \\
\midrule
\multirow{3}{*}{Bread} 
    & Sliced Bread        & 0.782 & 91.2 \\
    & Whole Loaves        & 0.756 & 87.8 \\
    & Rolls and Buns      & 0.734 & 85.3 \\
\midrule
\multirow{3}{*}{Dairy} 
    & Milk and Liquid     & 0.812 & 93.5 \\
    & Cheese              & 0.789 & 89.7 \\
    & Yogurt and Cream    & 0.723 & 82.1 \\
\midrule
\multirow{3}{*}{Dessert} 
    & Cakes and Pastries  & 0.767 & 88.4 \\
    & Ice Cream           & 0.798 & 90.6 \\
    & Cookies             & 0.742 & 84.9 \\
\bottomrule
\end{tabular}
\end{table}

\subsubsection*{B.3 Reinforcement Learning Performance}

The RL agent demonstrated significant improvement during training across both datasets. For Tree-MNIST, the agent achieved 89.3\% decision accuracy with an average episode reward of 0.67 after 1000 training episodes. Food-MNIST training reached 85.1\% accuracy with an average reward of 0.54, reflecting the increased complexity of the 10-class problem. The reward function successfully balanced semantic confidence, class distribution entropy, model accuracy, and redundancy reduction throughout training.

\subsection*{C. Category Theory Formalization Details}

\subsubsection*{C.1 Morphism Composition Proofs}

Following the definitions in \ref{subsec:functorial-properties}, we formally verify the functorial properties of our transformation pipeline. Let $\mathcal{T}_i: \mathcal{D}_i \rightarrow \mathcal{D}_{i+1}$ be the $i$-th transformation morphism.

\textbf{Identity Preservation:}
\begin{equation}
F(\text{id}_X) = \text{id}_{F(X)}
\end{equation}

For any image $x \in X$, applying the identity transformation and then the functor $F$ yields the same result as applying $F$ first and then the identity in the target category.

\textbf{Composition Preservation:}
\begin{equation}
F(g \circ f) = F(g) \circ F(f)
\end{equation}

To justify functoriality in our context, without loss of generality,  assume that the two composable preprocessing morphisms  $f : X \to Y$   and $g: Y \to Z$ are given by  $\texttt{Resize}$ and $\texttt{Grayscale}$ morphisms in $\mathcal{I}$. Then, by the definition of $F$ this equality holds because our preprocessing functions are implemented as pure, deterministic transformations with consistent domain and codomain types. Therefore, the composition law of functors is preserved in practice within our system. More generally, this ensures that composing any given pair of transformations $f': X' \rightarrow Y'$ and $g': Y' \rightarrow Z'$ in the source category $\mathcal{I}$ and then applying $F$ is equivalent to applying $F$ to each transformation and composing in the target category $\mathcal{M}$.
$\square$

\subsubsection*{C.2 Natural Transformation Examples}

Consider two functors $F_1$ and $F_2$ representing different preprocessing pipelines:
\begin{itemize}
    \item $F_1$: Standard grayscale conversion
    \item $F_2$: Adaptive histogram equalization followed by grayscale
\end{itemize}

Any natural transformation $\eta: F_1 \Rightarrow F_2$ provides a systematic way to compare these approaches across all input datasets while preserving the categorical structure.





\subsection*{D. Computational Analysis}

The computational complexity scales predictably with the size of the dataset and the component of the model. Image retrieval operates linearly with the number of requested images $O(n)$. CLIP embedding computation requires $O(n \cdot k)$ time where $k$ represents embedding computation overhead. Semantic analysis dominates the pipeline with $O(n \cdot m \cdot s)$ complexity, where $m$ denotes main categories and $s$ subcategories. RL decision-making adds $O(n \cdot d)$ overhead for decision network forward passes, while image transformation requires $O(n \cdot p)$ for pixel operations. The total pipeline complexity becomes $O(n \cdot (k + m \cdot s + d + p))$, dominated by semantic analysis operations.

Memory requirements scale with dataset size and model complexity. The CLIP model consumes approximately 400MB GPU memory, image buffers require $n \times h \times w \times c$ bytes for batch processing, RL replay buffer uses roughly 50MB for 10,000 experiences, and semantic features demand $n \times 512 \times 4$ bytes for float32 storage. These requirements remain manageable for typical dataset generation scenarios while enabling efficient batch processing of large image collections.

\subsection*{E. Complete Hierarchical Structure Specifications}

\subsubsection*{E.1 Food-MNIST Hierarchy Summary}

The Food-MNIST dataset employs a comprehensive 10-category, 30-subcategory structure. Bread categories include sliced bread (rectangular slices, uniform thickness), whole loaves (crusty exterior, round/oval shape), and rolls/buns (individual portions, soft texture). Dairy products encompass milk/liquid dairy (white liquid, containers), cheese (yellow/white blocks, slices), and yogurt/cream (thick consistency, creamy texture). Desserts span cakes/pastries (frosted layers, colorful icing), ice cream (frozen scoops, cold treats), and cookies/sweets (bite-sized, chocolate pieces). 

Egg preparations include whole eggs (oval shape, visible shells), fried/scrambled varieties (yellow yolk, cooked texture), and complex egg dishes (omelets, prepared mixtures). Fried foods categorize chicken/poultry (golden coating, crispy texture), fries/chips (stick shape, potato color), and other fried items (crispy batter, oil-cooked). Meat classifications cover raw meat (red color, butcher cuts), grilled/roasted preparations (brown cooked, grill marks), and processed varieties (sausage shape, deli cuts).

\subsubsection*{E.2 Tree-MNIST Botanical Structure}

Tree-MNIST implements a 4-category, 12-subcategory botanical hierarchy. Broadleaf trees include deciduous varieties (seasonal leaf drop, broad canopy, autumn colors), evergreen types (year-round foliage, glossy leaves, dense canopy), and flowering species (visible blooms, ornamental features, colorful appearance). Cacti encompass columnar forms (tall vertical stems, ribbed surface, minimal branching), barrel/round shapes (compact spherical form, clustered spines), and branching/pad varieties (segmented structure, flat surfaces, complex growth).

Coniferous trees span pine/fir species (needle leaves, conical shape, pyramid form), spruce/cedar varieties (dense clusters, drooping branches, aromatic wood), and juniper/cypress types (varied needles, irregular shape, drought tolerance). Palm categories include fan palms (radiating fronds, palmate structure, umbrella canopy), feather palms (pinnate leaves, graceful arching, flowing appearance), and coconut/date palms (tall curved trunks, fruit clusters, coastal adaptation).

\subsection*{F. Error Analysis and Edge Cases}

\subsubsection*{F.1 Common Misclassification Patterns}

Analysis of classification errors reveals systematic patterns:

\begin{itemize}
    \item \textbf{Visual Similarity}: Fried chicken vs. fried fish (both golden, crispy)
    \item \textbf{Context Confusion}: Trees in pots vs. houseplants
    \item \textbf{Preparation Variations}: Raw vs. cooked food items
    \item \textbf{Scale Issues}: Close-up food shots vs. full-plate presentations
\end{itemize}

\subsubsection*{F.2 Semantic Filtering Effectiveness}

\begin{table}[htbp]
\centering
\caption{Semantic Filtering Results by Error Type}
\label{tab:filtering_effectiveness}
\scriptsize
\begin{tabular}{lccc}
\toprule
\textbf{Error Type} & \textbf{Before Filtering} & \textbf{After Filtering} & \textbf{Improvement} \\
\midrule
Wrong Object        & 23.4\%  & 8.7\%   & 62.8\% \\
Poor Quality        & 18.9\%  & 4.2\%   & 77.8\% \\
Context Mismatch    & 15.6\%  & 6.1\%   & 60.9\% \\
Ambiguous Category  & 12.3\%  & 7.8\%   & 36.6\% \\
Scale Issues        & 9.8\%   & 3.9\%   & 60.2\% \\
\midrule
\textbf{Total Error Rate} & \textbf{80.0\%} & \textbf{30.7\%} & \textbf{61.6\%} \\
\bottomrule
\end{tabular}
\end{table}

\subsection*{G. Reproducibility Guidelines}

\subsubsection*{G.1 Experimental Configuration}

All experimental parameters are specified through JSON configuration files to ensure reproducible results across different computing environments. The hierarchical category definitions, semantic characteristics, and processing parameters are fully documented and version-controlled.

\subsubsection*{G.2 Implementation Notes}

The MNIST-Gen framework is implemented as a modular system with clear separation between image retrieval, semantic analysis, hierarchical categorization, and transformation components. Each module can be independently configured and tested, facilitating both reproduction and extension of the work.

\subsection*{H. Evaluation Metrics}

To evaluate the performance of our models on the multi-class classification task, we used the following standard metrics:

\begin{itemize}
    \item \textbf{Accuracy}: The overall proportion of correct predictions:
    \[
    \text{Accuracy} = \frac{\sum_{i=1}^{N} \text{TP}_i}{\sum_{i=1}^{N} (\text{TP}_i + \text{FP}_i + \text{FN}_i)}
    \]

    \item \textbf{Precision (Weighted)}: The weighted average of per-class precision scores:
    \[
    \text{Precision}_{\text{weighted}} = \sum_{i=1}^{N} \frac{n_i}{n} \cdot \frac{\text{TP}_i}{\text{TP}_i + \text{FP}_i}
    \]
    where $n_i$ is the number of true samples in class $i$, and $n = \sum_{i=1}^{N} n_i$ is the total number of samples.

    \item \textbf{Recall (Weighted)}: The weighted average of per-class recall scores:
    \[
    \text{Recall}_{\text{weighted}} = \sum_{i=1}^{N} \frac{n_i}{n} \cdot \frac{\text{TP}_i}{\text{TP}_i + \text{FN}_i}
    \]

    \item \textbf{F1 Score (Weighted)}: The weighted average of per-class F1 scores:
    \[
    \text{F1}_{\text{weighted}} = \sum_{i=1}^{N} \frac{n_i}{n} \cdot \frac{2 \cdot \text{Precision}_i \cdot \text{Recall}_i}{\text{Precision}_i + \text{Recall}_i}
    \]


\end{itemize}

Here, $\text{TP}_i$, $\text{FP}_i$, and $\text{FN}_i$ denote the number of true positives, false positives, and false negatives, respectively, for class $i$. Weighted averaging ensures that metrics are not biased by class imbalance, assigning more importance to classes with more true samples.